\documentclass[11pt]{article}
\usepackage{coling2020}
\usepackage{times}
\usepackage{latexsym}
\usepackage{amsmath,amsfonts,bm}
\usepackage{xcolor}
\usepackage{arydshln}
\usepackage{graphicx}
\usepackage[tight]{subfigure}
\usepackage[T1]{fontenc}
\usepackage[hyphens]{url}
\usepackage{enumitem}
\usepackage[toc,title,page]{appendix}
\usepackage{booktabs}

\colingfinalcopy 

\title{End to End Binarized Neural Networks for Text Classification}

\author{Harshil Jain$^{1}$\thanks{\hspace{2mm}The authors contributed equally to this research and work done at NeuralSpace}  , Akshat Agarwal$^{2}$\footnotemark[1] , Kumar Shridhar$^{3}$\footnotemark[1] , Denis Kleyko$^{4,5}$ \\
  $^{1}$Computer Science and Engineering, IIT Gandhinagar, Gujarat, India \\
  $^{2}$Electrical Engineering, Delhi Technological University, Delhi, India \\
  $^{3}$NeuralSpace, London \\
  $^{4}$Redwood Center for Theoretical Neuroscience, University of California, Berkeley\\
  $^{5}$Intelligent Systems Lab, Research Institutes of Sweden\\
  {\tt jain.harshil@iitgn.ac.in}, 
  {\tt akshat.agarwal0311@gmail.com}\\
  {\tt kumar@neuralspace.ai},
  {\tt denis.kleyko@ri.se} \\}

\begin{document}

\maketitle
\begin{abstract}

Deep neural networks have demonstrated their superior performance in almost every Natural Language Processing task, however, their increasing complexity raises concerns. 
In particular, these networks require high expenses on computational hardware and training budget is a concern for many. 
Even for a trained network, the inference phase can be too demanding for resource-constrained devices, thus limiting its applicability. 
The state-of-the-art transformer models are a vivid example.
Simplifying the computations performed by a network is one way of relaxing the complexity requirements. 
In this paper, we propose an end to end binarized neural network architecture for the intent classification task. 
In order to fully utilize the potential of end to end binarization, both input representations (vector embeddings of tokens statistics) and the classifier are binarized. 
We demonstrate the efficiency of such architecture on the intent classification of short texts over three datasets and for text classification with a larger dataset. 
The proposed architecture achieves comparable to the state-of-the-art results on standard intent classification datasets while utilizing $\sim$ $20$-$40$\% lesser memory and training time. Furthermore, the individual components of the architecture, such as binarized vector embeddings of documents or binarized classifiers, can be used separately with not necessarily fully binary architectures.
\end{abstract}

\section{Introduction}
\label{intro}

%
%
\blfootnote{
    \hspace{-0.65cm}  
    This work is licensed under a Creative Commons 
    Attribution 4.0 International License.
    License details:
    \url{http://creativecommons.org/licenses/by/4.0/}}
In recent years, deep neural networks have achieved great success in a variety of domains including but not limited to autonomous driving \cite{grigorescu2020survey}, image recognition \cite{krizhevsky2012imagenet}, and text classification \cite{kim2014convolutional}.
However, deep networks are becoming more and more computationally expensive due to the ever-growing size of the models being trained.
This tendency has been noticed in~\cite{EnergyNLP} and it has been recommended that academia and industry researchers should draw their attention towards more computationally efficient methods. 
At the same time, many important application areas such as chatbots, IoT devices, mobile devices and other types of power-constrained and resource-constrained platforms require solutions that would be  highly computation and memory efficient. 
Such use-cases limit the potential use of the state-of-the-art deep networks.
One viable solution is the transformation of these high-performance neural networks to a more computationally efficient architecture. 
For example, continuous values of the network's parameters (e.g.,  float $32$-$64$ bits) can be quantized into discrete integer values. 
Recently, Binarized Convolutional Neural Networks (BNN)~\cite{courbariaux2016binarized} have been developed where both weights and activations are restricted to $\{+1, -1\}$. 
This results in highly computationally efficient networks with a much lower memory footprint. 
However, to the best of our knowledge, in the area of text classification in general and intent classification, in particular,  no end to end trainable binarized architecture has been demonstrated yet. 

In this paper, we introduce an architecture for the task of intent classification that fully utilizes the power of binary representations. 
The input representations are tokenized and embedded in binary high-dimensional (HD) vectors forming distributed representations
using the approach known as hyperdimensional computing~\cite{Kanerva09}, \cite{Rachkovskij2001}. 
The binary input representations are used for training an end to end BNN classifier for intent classification. 
Classification performance-wise, the binarized architecture achieves the results comparable to the state-of-the-art on three standard intent classification datasets~\cite{braun-EtAl:2017:SIGDIAL} but performs less accurately on a large text classification 20NewsGroups~\cite{Lang95} dataset.
The efficiency of the proposed architecture is shown in terms of its time and memory complexity relative to non-binarized architectures. 
For example, it reduces the memory consumption by about $40$\% when compared to a non-binarized neural network using GloVe embeddings~\cite{pennington2014glove} as the input representation. 

\noindent Thus, the main contributions of this paper are as follows: 
\begin{itemize}[leftmargin = *]
    \itemsep-0.25em 
    \item We introduce a fully end to end binarized architecture for the task of intent classification that uses binary representations followed by the binarized classifier.
    \item We show that the proposed architecture preserves classification performance while being more time and memory-efficient compared to the non-binarized architecture. 
    \item Finally, we show that the components of the architecture can be used individually as parts of solutions for other NLP tasks.
\end{itemize}

\begin{figure}
\centering
\includegraphics[width=1\linewidth]{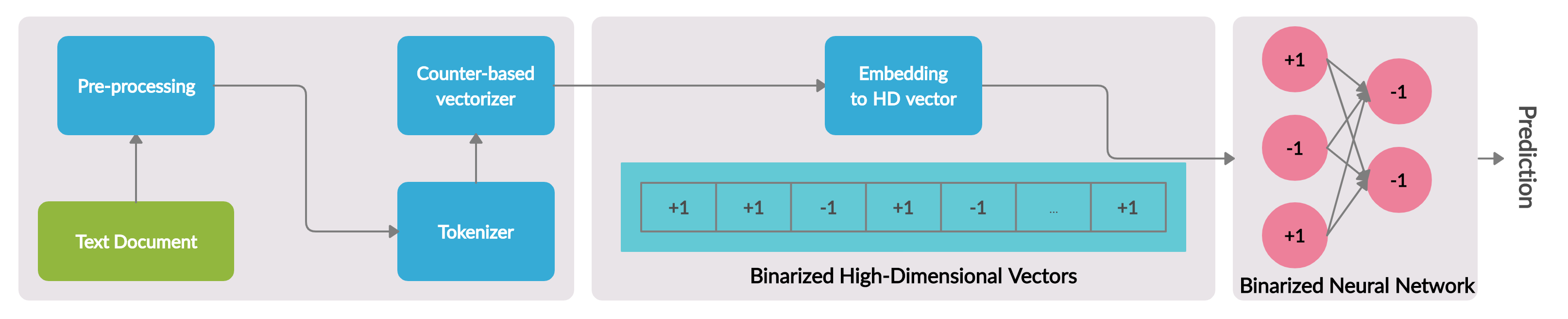}
\caption{A schematic diagram of the end to end binarized classification architecture for text classification}
\label{fig:mainImage}
\end{figure}

\section{Related Work}
\label{sec:related}

\subsection{Tokenization}
Classical methods such as word-based tokenizers were one of the most popular tokenization methods to form a dictionary. 
They, however, suffer from out-of-dictionary problems for unseen words~\cite{bojanowski2017enriching}.
$n$-gram based tokenization estimates the out-of-dictionary words better.
They can also be used to represent words~\cite{2016arXiv160701759J}. 
As an alternative to word/token representation~\cite{SemHash19} introduced the idea of Subword Semantic Hashing that marks the start and end of a word with a specialized token and defines a hashing function to form $n$-gram-based subword tokens. Byte-Pair-Encoding (BPE)~\cite{sennrich2015neural} is another popular subword segmentation method widely used in many neural machine translation systems. The main idea of the BPE is to split the sentence into individual tokens and to merge the most frequent tokens to get the final representation. The area was further extended by
~\cite{kudo2018subword} using a unigram language model. 
Most recently, a highly successful Bidirectional Encoder Representations from Transformers (BERT) model~\cite{devlin2018bert} has proven to be a good tokenization choice where the tokenization is based on BERT's masking objective. 

\subsection{Representation of tokens}
Conventional $n$-gram statistics uses sparse localist representations.
In the special case, when $n=1$, it can be seen as count-based vectors and compatible with Term Frequency Inverse Document Frequency (TF-IDF).
Note that these representations are proportional to the length of the tokens' dictionary. 
Moreover, the length of representations grows exponentially with $n$, making them highly inefficient in terms of memory. 
Hyperdimensional computing-based embedding of $n$-gram statistics~\cite{RIJHK2015},~\cite{alonso2020hyperembed} is used to represent the tokens extracted from the input text into a vector of fixed size, which does not depend on $n$. 
Previous works have shown the usefulness of these embeddings for the tasks of language identification~\cite{RahimiLPHD}, \cite{PSI19} and news articles classification~\cite{Rasti2016}. 
This work differs from the ones mentioned above, as in these studies, the embeddings were not combined with a trainable architecture. 
In contrast, here, the binary embeddings are used as a part of the architecture representing the input text, which enters a BNN. 


As an alternative to $n$-gram based representations, dense trainable word embeddings (e.g., word2vec~\cite{mikolov2013efficient} or  GloVe~\cite{pennington2014glove}) use a fixed size vector representation. 
These embeddings are based on the distributional hypothesis and are expected to represent words with similar meanings nearby in the embedding space. 
However, the drawbacks of such embeddings are that they require pre-training on large text corpora and that storage of the trained embeddings demands a non-negligible memory footprint.

\subsection{Classification models}
In recent years, convolutional neural networks have been shown to be useful for NLP classification tasks at the level of sentences~\cite{kim2014convolutional} when trained on top of the pre-trained word embeddings. 
Moreover, token-level convolutional networks~\cite{NIPS2015_5782} have achieved state-of-the-art or competitive results on several large scale classification datasets. 
BNNs~\cite{courbariaux2016binarized}, on the other hand, have been extensively used for computer vision tasks and their application to NLP tasks has been limited. 
One such example is~\cite{zhengbinarized}, where the binarized recurrent neural network has been used for the language modeling task.  
In this work, we show that BNNs are applicable for intent classification. 
We further show that the  binarization capabilities are enhanced when using a fully binarized architecture, including embeddings and classifiers.

\section{Methods}
\label{sec:methods}

This section introduces a variety of methods used in the experiments. 
Figure~\ref{fig:mainImage} presents a schematic overview of the architecture. 
Given an input text document $D$, we first pre-process the document (see Section~\ref{sect:exp:det} for more details). 
The pre-processed document is then tokenized into the corresponding tokens $<T_1, T_2, ... , T_n>$ using one of the methods below. 
The tokens are used as an input to a count-based vectorizer. 
The representation of vectorizers, which is sparse and localist, is embedded into an HD vector (distributed representation) using hyperdimensional computing~\cite{HDNP17}. 
Hyperdimensional computing is tightly related to randomly connected neural networks~\cite{DensityEncoding2020}, \cite{intESN2017} but it allows constructing more complex representations such as data structures~\cite{HD_FSA}. 
Moreover, HD vector representing the counter's content can be binarized. 
It is used as an input to a classifier.
The main classifier studied in this work is BNN but other classifiers are also considered for benchmarking. 



\subsection{Tokenizers}
\subsubsection{Word tokenizer}

Given a pre-processed text $D$, a word tokenizer splits the input text $D$ into tokes $<T_1, T_2, ... , T_n>$ corresponding to words. Words are split based on spaces while throwing away certain tokens, such as punctuation marks, unidentified tokens, hyphen symbols and others. 
For a sample text sentence $D$, [\textit{``hello! how are you?''}], its word tokens would be [\textit{``hello'', ``how'', ``are'', ``you''}].

\subsubsection{Semantic Hashing tokenizer}

First, Subword Semantic Hashing (SemHash) splits the input text $D$ into word tokens $<T_1, T_2, ... , T_n>$. 
Next, for each word token $T_i$, SemHash creates subword tokens using a hashing method.  
For a given pre-processed input sample text $D$, e.g., [\textit{``hello how are you''}], the output of the word split-step would be: [\textit{``hello'', ``how'', ``are'', ``you''}].
A hashing function $\mathcal{H}(T_i)$ adds $\#$ at the beginning and end of each word token. 
Subwords are extracted for every token as $n$-grams for a given value of $n$, for example, when $n=3$, $\mathcal{H}(hello)$=[\textit{``\#he'', ``hel'', ``ell'', ``llo'', ``lo\#''}].
 
\subsubsection{Byte Pair Encoding tokenizer}
Given as input a pre-processed text $D$, BPE \cite{sennrich2015neural} splits the sentences into individual words with special end of word token $\dot{g}$ and merges the most frequent tokens to get the final tokens. This is highly efficient for representing the unknown words into known frequent token sets. 
For a given input sample text $D$, e.g., [\textit{``hello how are you''}], the output of the BPE tokenizer would be: 
[\textit{``hello$\dot{g}$'', ``how$\dot{g}$'', ``are$\dot{g}$'', ``you''}].

\subsubsection{Char level Byte Pair Encoding tokenizer}
For better subword sampling, \cite{kudo2018subword} proposed a new subword segmentation method based on a unigram language model. The unigram language model is based on an independent subword occurrence assumption with the probability of a subword sequence defined as the product of the subword occurrence probabilities.
For a given input sample text $D$, e.g., [\textit{``hello how are you''}], the output of the Char level BPE tokenizer would be: 
[\textit{``h'', ``e'', ``l'', ``l'', ``o$\dot{g}$'', ``h'', ``o'', ``w$\dot{g}$'', ``a'', ``r'', ``e$\dot{g}$'', ``y'', ``o'', ``u$\dot{g}$''}] 
where $\dot{g}$ is the special token to mark the end of the word. 

\subsubsection{BERT-based tokenizer}
The BERT-based tokenizer is based on the BERT architecture, as proposed by \cite{devlin2018bert}. It learns the token splits based on the masking objective on an unsupervised dataset. 
For a given input sample text $D$, e.g., [\textit{``hello how are you''}], the output of the BERT-based tokenizer would be: [\textit{``[cls]'', ``hello'', ``how'', ``are'', ``you'', ``[sep]''}] where ``[cls]'' is the special token used to mark the start of the sentence and ``[sep]'' is the sentence separator. 

%



\begin{figure}[t]
\hspace{10 mm} \subfigure[t-SNE projection with GloVe embeddings]{\includegraphics[width=0.38\linewidth]{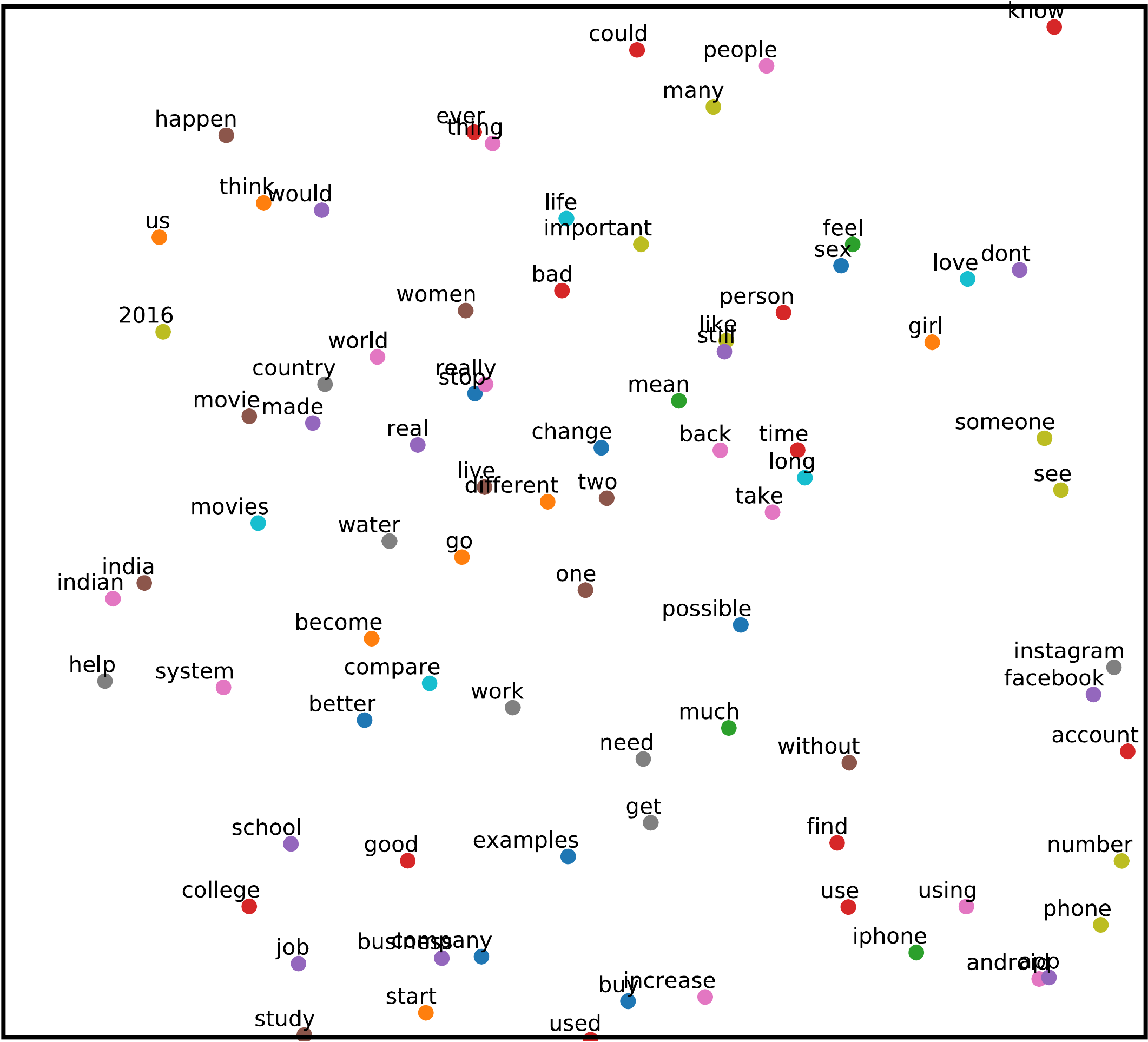}}\hspace{10 mm}
\subfigure[t-SNE projection with HD embeddings]{\includegraphics[width=0.41\linewidth]{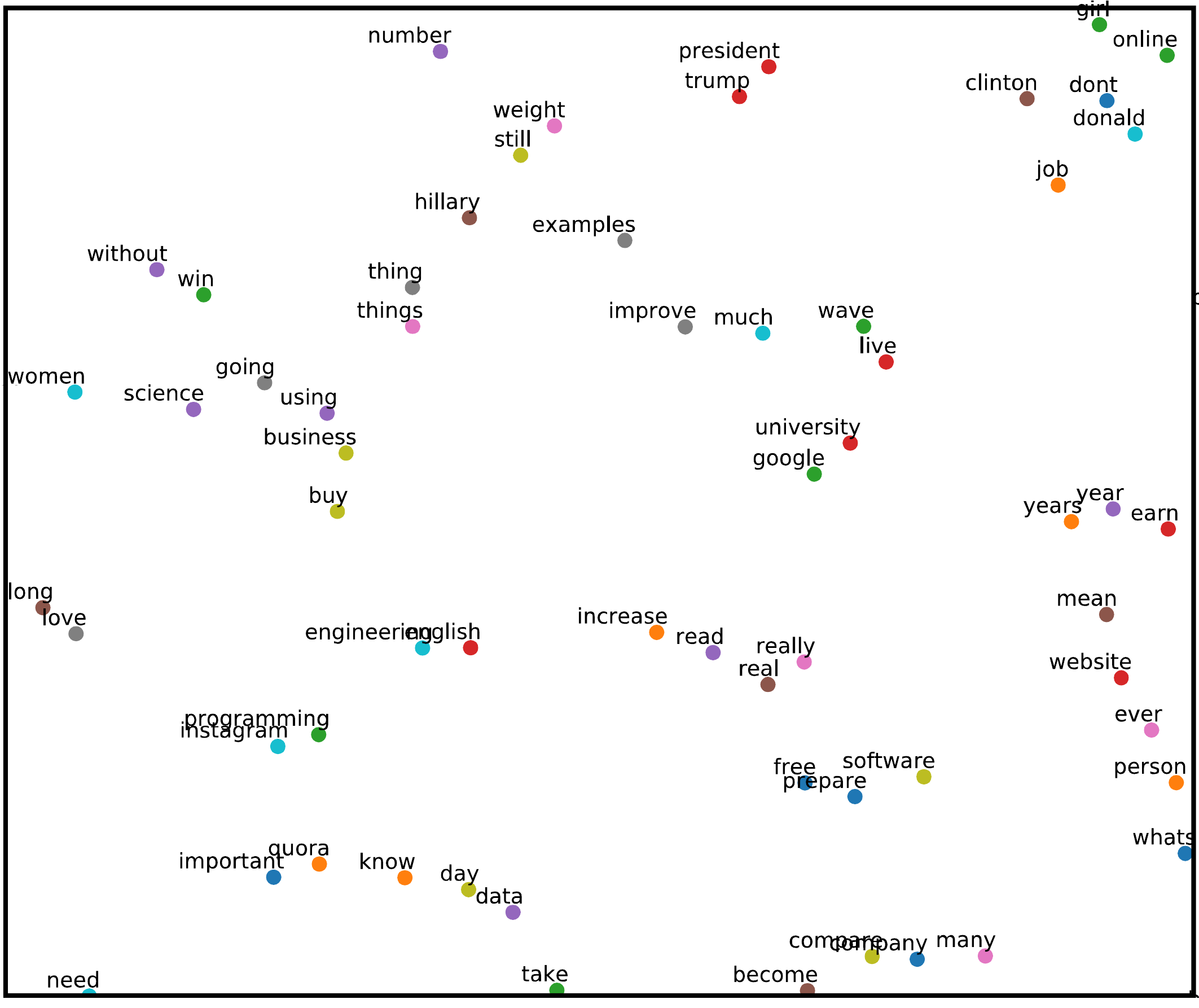}}\hfill

\caption{(a) shows the t-SNE projection with GloVe embeddings, while (b) shows the t-SNE projection with HD vectors; GloVe is pre-trained on the Quora question pairs dataset\protect \footnotemark   while HD vectors are directly obtained from $n$-gram statistics of the words present in the plot.
Colours of points are not associated to any meaning. 
}
\label{fig:tSNE-HDvsword2vec}
\end{figure}
\footnotetext{\url{https://www.kaggle.com/c/quora-question-pairs}}

\subsection{High-Dimensional embedding of vectorized representations}
\label{sec:HD:embed}



The vectorized representations containing the statistics of tokens extracted from the pre-processed documents are usually sparse and have high dimensionality.
In order to reduce the dimensionality of representations, we use hyperdimensional computing~\cite{FradySDR2020}. 
First, each unique token $T_i$ is assigned with a random $d$-dimensional bipolar HD vector, where $d$ would be a hyperparameter of the method. 
HD vectors are stored in the item memory, which is a matrix $\textbf{H} \in [d \times n]$.
Thus, for a token $T_i$ there is an HD vector $\textbf{H}_{T_i} \in \{-1, +1\}^{[d \times 1]}$. 
To construct composite representations from the basic HD vectors stored in $\textbf{H}$, hyperdimensional computing defines three key operations: permutation ($\rho$),  
binding ($\odot$, implemented via position-wise multiplication), and
bundling ($+$, implemented via position-wise addition).
For detailed properties and definitions of hyperdimensional computing operations, please see~\cite{Kanerva09}.
The bundling operation allows storing information in HD vectors~\cite{Frady17}.
Three operations above allow embedding vectorized representations based on $n$-gram statistics into an HD vector~\cite{RIJHK2015}. 
Note that a vectorizer that simply counts the number of times each token is present in the document is the special case of $n$-gram statistics for $n=1$.

We first generate $\textbf{H}$, which has an HD vector for each token.
The permutation operation $\rho$ is applied to  $\textbf{H}_{T_j}$ $j$ times ($\rho^j (\textbf{H}_{T_j})$) to represent a relative position of token $T_j$ in an $n$-gram.
A single HD vector corresponding to an $n$-gram (denoted as $\textbf{m}$) is formed using the consecutive binding of permuted HD vectors $\rho^j (\textbf{H}_{T_j})$ representing tokens in each position $j$ of the $n$-gram. 
For example, the trigram  `\#he' will be embedded to an HD vector as follows: $\rho^{1}(\textbf{H}_{\text{\#}}) \odot  \rho^{2}(\textbf{H}_{\text{h}}) \odot \rho^{3}(\textbf{H}_{\text{e}}) $. In general, the process of forming HD vector of an $n$-gram is formalized as follows: 
\noindent
\begin{equation*}
 \textbf{m} = \prod_{j=1}^{n} \rho^{j}(H_{T_j}), 
\end{equation*}
\noindent
where $T_j$ is token in $j$th position of the $n$-gram;
the consecutive binding operations applied to $n$ HD vectors are denoted by $\prod$.
Once it is known how to form an HD vector for an individual $n$-gram, embedding the $n$-gram statistics into an HD vector $\textbf{h}$ is achieved  by bundling together all $n$-grams observed in the document:
\noindent
\begin{equation*}
\textbf{h}= \lbrack
 \sum_{i=1}^{k} f_{i} \textbf{m}_i = \sum_{i=1}^{k} f_i   \prod_{j=1}^{n} \rho^{j}(H_{T_j}) \rbrack, 
\end{equation*}
\noindent
where $k$ is the total number of $n$-grams; $f_i$ is the frequency of $i$th $n$-gram and $\textbf{m}_i$ is the HD vector of $i$th $n$-gram;
$\sum$ denotes the bundling operation when applied to several HD vectors; $[*]$ denotes the binarization operation, which is implemented via the sign function.
The usage of $[*]$ is optional, so we can either obtain binarized or non-binarized $\textbf{h}$.
If $\textbf{h}$ is non-binarized, its components will be integers in the range $[-k,k]$ but these extreme values are highly unlikely since  HD vectors for different $n$-grams are quasi-orthogonal, which means that in the simplest (but not practical) case when all $n$-grams have the same probability the expected value of a component in $\textbf{h}$ is $0$. 
Due to the use of $\sum$ for representing $n$-gram statistics, two HD vectors embedding two different $n$-gram statistics might have very different amplitudes if the frequencies in these statistics are very different. 
When HD vectors $\textbf{h}$ are binarized, this issue is addressed. In the case of non-binarized HD vectors, we address it by using the cosine similarity, which is imposed by normalizing each  $\textbf{h}$ by its $\ell _2$ norm, thus, all $\textbf{h}$ have the same norm and their dot product is equivalent to their cosine  similarity.
Figure~\ref{fig:tSNE-HDvsword2vec} provides an example of t-SNE projections of words represented either as GloVe embeddings or as embedding of $n$-gram statistics to HD vectors.  

\subsection{Binarized Neural Networks}

Based on the work of~\cite{courbariaux2016binarized}, we construct BNNs capable of working with representations of texts. To take the full advantage of binarized HD vectors, we constraint the weights and activations of the network layers to be $\{+1, -1\}$. 
This constraint is highly efficient in terms of hardware and memory, as bit-wise operations are used instead of multiply-accumulate operations. For example, a multiplication on binary values can be performed using an XNOR logical operation.

The vectorized representations of tokens embedded into HD vectors are binarized with all values $\{+1, -1\}$. 
In the case of HD vectors, we binarize the result of the bundling operation using the ${\rm sign}$ function. 
Similarly, the ${\rm sign}$ function is used in the BNN for every weight or activation to restrict them into  $\{+1, -1\}$ as follows:
\begin{equation}
    b(x) = [x]= {\rm sign}(x) = \left\{ \begin{array}{ll}
                        +1 & \mbox{if $x \geq 0$},\\
                        -1 & \mbox{otherwise}\end{array} \right.
\end{equation}
where, $x$ can be any weight or activation value. 

We further define a convolutional 1D layer that creates a convolution kernel that is convolved with the input HD vector over a single spatial dimension to produce a tensor of outputs. 
Since gradient descent methods make small changes to the value of the weights, which cannot be done with binary values, we use the straight-through estimator (STE) idea, as mentioned in~\cite{yin2019understanding}.
We also define a value over which we clip the gradients in the backward pass:
\begin{equation}
    \frac{\delta b(x)}{\delta x} = \left\{ \begin{array}{ll}
                        +1 & \mbox{if $|x| < \text{clip value}$},\\
                        0 & \mbox{otherwise}\end{array} \right.
\end{equation}
This ensures that the entire architecture is end to end trainable using the gradient descent optimization. 

\section{Empirical Analysis}
\subsection{Datasets}

All the experiments in this paper are performed on four datasets, namely: the \textit{Chatbot Corpus} (Chatbot), the \textit{Ask Ubuntu Corpus} (AskUbuntu), the \textit{Web Applications Corpus} (WebApplication), and the \textit{20 News Groups Corpus} (20NewsGroups). The Chatbot, AskUbuntu and WebApplication datasets are used for short text intent classification and were taken from~\cite{braun-EtAl:2017:SIGDIAL}. The datasets are available on GitHub\footnote{Under the Creative Commons CC BY-SA 3.0 license: \url{https://github.com/sebischair/NLU-Evaluation-Corpora}}; 20NewsGroups dataset is available within the sklearn \cite{sklearn_api} library. 

\paragraph{Chatbot Corpus:} The Chatbot Corpus was collected from questions posted using a Telegram Chatbot. The classification task was performed on $2$ intents: \textit{(Departure Time and Find Connection)} with $206$ questions. The train set consists of $100$ samples, while $106$ samples were used as the test set. 

\paragraph{AskUbuntu Corpus:} The AskUbuntu Corpus has been extracted from the AskUbuntu platform. It consists of $5$ intents \textit{(Make Update, Setup Printer, Shutdown Computer, Software Recommendation, and None)} with $53$ training samples and $109$ test samples. 

\paragraph{WebApplication Corpus:} The WebApplication Corpus consists of $88$ samples and $8$ intents \textit{(Change Password, Delete Account, Download Video, Export Data, Filter Spam, Find Alternative, Sync Accounts, and None)} with $30$ training samples and $58$ test samples. 

\paragraph{20NewsGroups Corpus:} The 20NewsGroups consists of news post categorized into $20$ categories. Each category has exactly $18846$ text samples with $11314$ samples for training and $7532$ samples for testing.
A detailed data distribution for each dataset is provided in the Appendix.

\begin{table}[t]

\begin{center}
\scalebox{0.90}{
\begin{tabular}{|c|c|c||c|c||c|c||c|c|} 
\cline{2-9}
\multicolumn{1}{c}{} & \multicolumn{2}{|c||}{{Chatbot}} & \multicolumn{2}{|c||}{{AskUbuntu}} & \multicolumn{2}{|c||}{{WebApplication}} & \multicolumn{2}{|c|}{{20NewsGroups}} \\ \hline

\hline
Tokenizers & \textit{Text-LeNet} & BNN & \textit{Text-LeNet} & BNN & \textit{Text-LeNet} & BNN & \textit{Text-LeNet} & BNN \\ \hline 
Word                 & \textbf{0.80} & 0.73 & 0.51 & \textbf{0.79}  & 0.56 & \textbf{0.78} & 0.54 & \textbf{0.56}   \\ \hdashline 
SemHash       & \textbf{0.94} & 0.90 & \textbf{0.87} & 0.84  & 0.79 & \textbf{0.83} & \textbf{0.78} & 0.69    \\ \hdashline 
BPE      & \textbf{0.80} & 0.58 & 0.54 & \textbf{0.67}  & 0.52 & \textbf{0.75} & 0.38 & \textbf{0.42}   \\ \hdashline 
Char BPE    & \textbf{0.92} & 0.81 & \textbf{0.76} & \textbf{0.76}  & \textbf{0.55} & 0.53 & \textbf{0.55} & 0.48    \\ \hdashline  
SentencePiece      & 0.80 & \textbf{0.99} & 0.70 & \textbf{0.72}  & 0.50 & \textbf{0.70} & 0.41 & \textbf{0.43}   \\ \hdashline 
BERT    & \textbf{0.89} & 0.88 & \textbf{0.72} & 0.71 & 0.70 & \textbf{0.77} & \textbf{0.60} & \textbf{0.60}    \\  
\hline
\end{tabular}}

\caption{ $F_1$ performance comparison of binarized \textit{Text-LeNet} (BNN) architecture with non-binarized \textit{Text-LeNet} for the task of intent classification on various datasets.}
\label{tab:classifiers:binarized}
\end{center}

\end{table}

\begin{table}[t]

\begin{center}
\scalebox{0.90}{
\begin{tabular}{|c|c|c|c|} 

\hline
Datasets & Binarized GloVe & Binarized SemHash & Binarized HD vectors  \\ \hline 
Chatbot                 & 0.74 & 0.91 & \textbf{0.99} \\ \hdashline 
AskUbuntu       & 0.86 & \textbf{0.87} & 0.84   \\ \hdashline 
WebApplication      & 0.66 & 0.80 & \textbf{0.83}   \\ \hdashline 
20NewsGroups     & 0.62  & 0.64 & \textbf{0.69}   \\   
\hline
\end{tabular}}
\caption{ $F_1$ performance comparison of Binarized GloVe vectors, Binarized SemHash vectors and Binarized HD vectors. All vectorizers use the same binarized \textit{Text-LeNet} architecture as classifier.}
\label{tab:classifiers_BNN}
\end{center}

\end{table}

\begin{figure}[t]
\begin{subfigure}[Memory Comparison]{\includegraphics[width=0.53\linewidth]{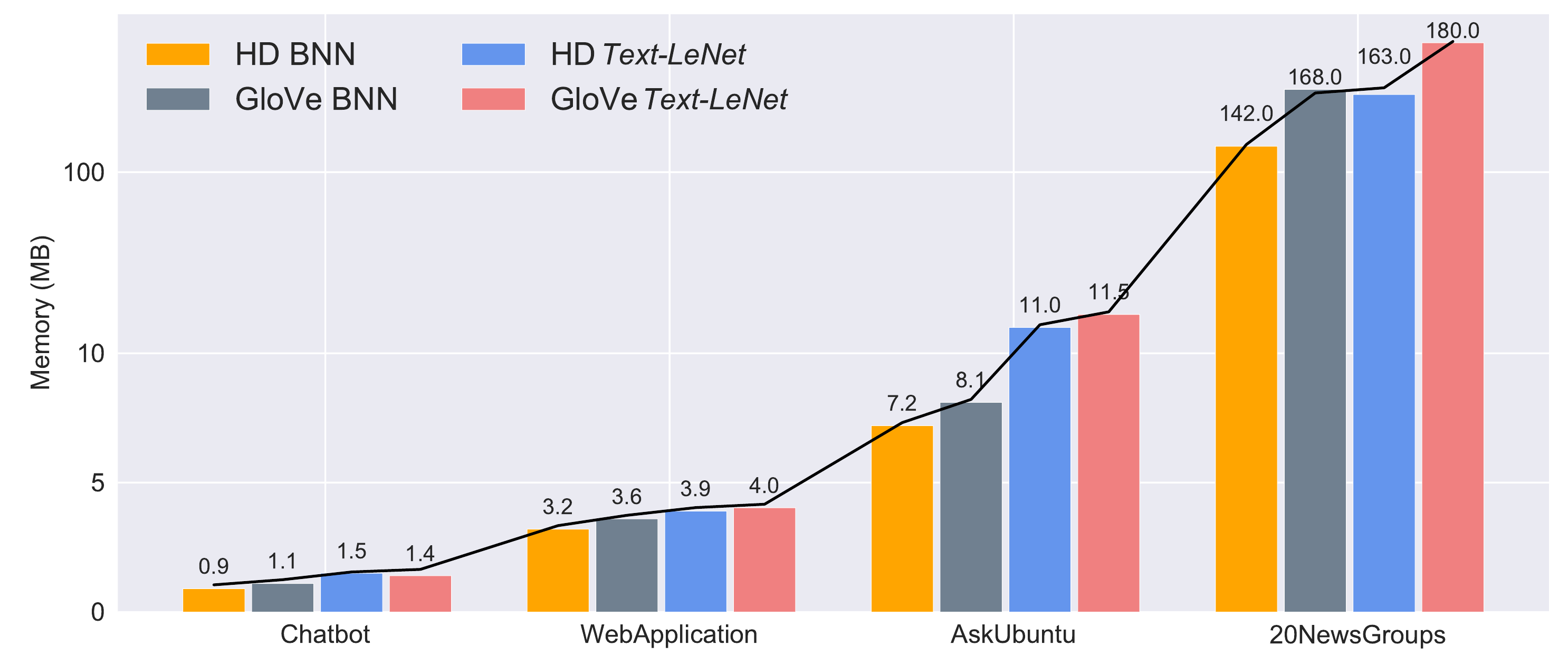}}\label{fig:memoryimage}\hfill
\end{subfigure}
\begin{subfigure}[Train Time per epoch Comparison]{\includegraphics[width=0.45\linewidth]{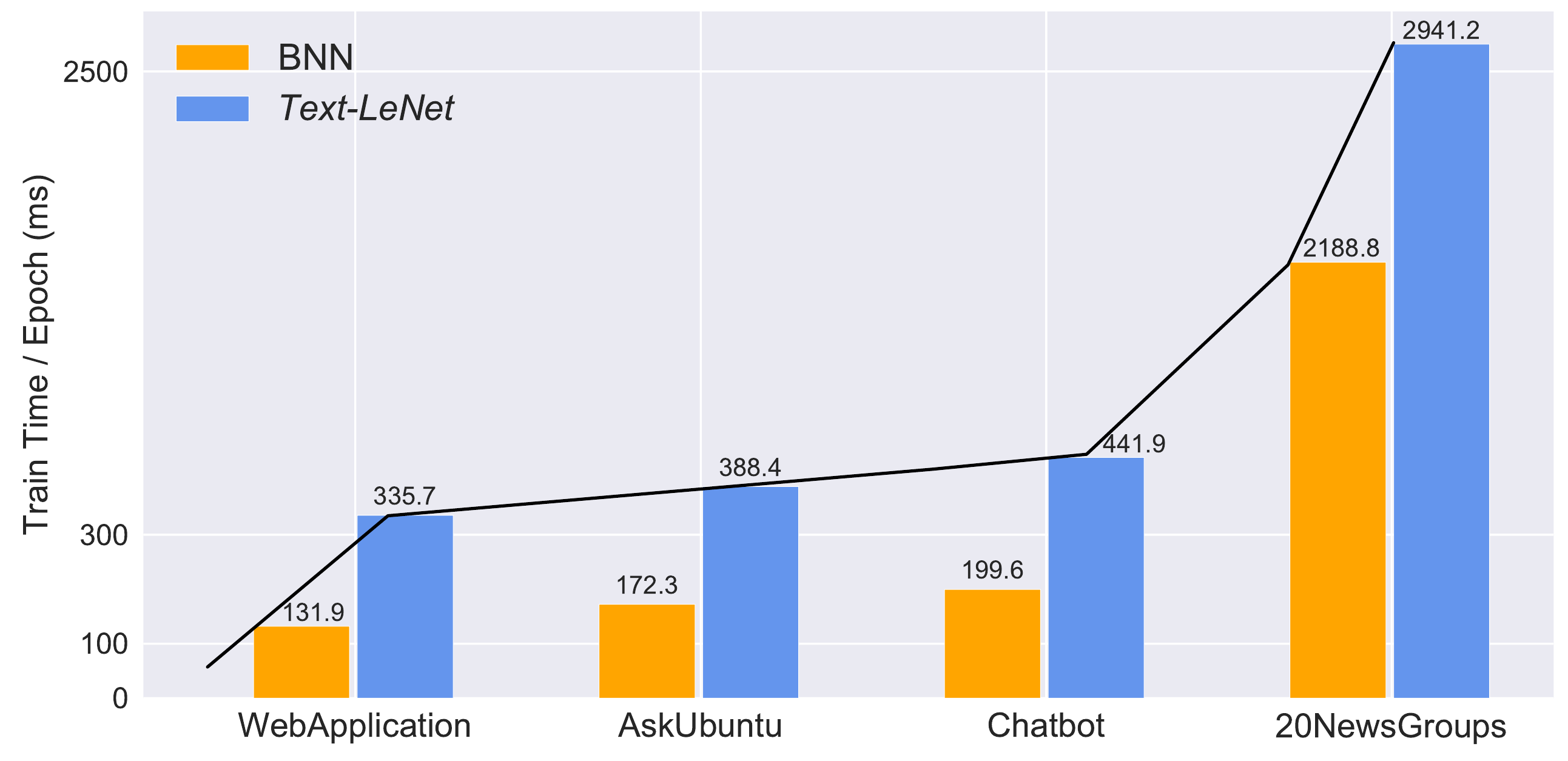}}\label{fig:timeimage}\hfill
\end{subfigure}
\caption{(a) shows the memory comparisons for all 4 datasets using HD \textit{Text-LeNet}, HD BNN, GloVe \textit{Text-LeNet}, GloVe BNN and (b) shows the training time per epoch comparison for all 4 datasets using BNN and \textit{Text-LeNet}}
\label{fig:memoryandtimeimage}
\end{figure}

\subsection{Experimental Details}
\label{sect:exp:det}

On all datasets, six tokenization methods were used, namely: Word-based, SemHash, BPE, Char level BPE, Sentence Piece and BERT-based. 
For Word-based tokenizer, the datasets were pre-processed using the Spacy library to remove stop words from data. Spacy pre-trained model ``en\_core\_web\_lg'' was used to parse the datasets. 
All text samples were also pre-processed by removing control tokens, in particular, the ones in the set \textit{[Cc]}, which includes Unicode tokens from \textit{U+0000} to \textit{U+009F}.
The small datasets (Chatbot, WebApplication and AskUbuntu) data samples within each class were made equal to the largest class sample size by augmenting the data. 
For SemHash, $n$ value was kept at $3$, and all trigrams were considered. For BPE, a dictionary size of $1000$ was used for Chatbot, WebApplication and AskUbuntu, and $2000$ for 20NewsCorpus dataset because of its larger size.  Similar values were used for Char-based BPE and Sentence Piece tokenizers. The difference between Sentence Piece and BPE is that Sentence Piece uses stop words removal before tokenization while BPE and Char-based BPE do not. BERT-based tokenizer was pre-trained on a large corpus  ``bert-large-cased-vocab.txt''\footnote{ \url{https://s3.amazonaws.com/models.huggingface.co/bert/bert-large-cased-vocab.txt}} that contains $29213$ unique words and has been provided by the Huggingface\footnote{\url{https://github.com/huggingface/}} library.

For the purpose of benchmarking, nine sklearn classifiers were applied to the intent classification datasets: MLP, Random Forest, Linear SVC, Passive Aggressive, SGD Classifier\footnote{The SGD classifier here refers to the SVM classifier trained using SGD optimization as per the sklearn library and this notation is used henceforth.}, Ridge Classifier, Nearest Centroid, Bernoulli NB, KNN Classifier with an HD vector size of  $d=\{512,1024,4098,8192,16384\}$. 
The detailed results for these classifiers are provided in the Appendix.
CNN-based architecture used $d=\{512,1024,4098,8192\}$ as embedding sizes. 
20NewsGroups dataset was run with MLP, Random Forest, Linear SVC and SGD classifiers for embedding sizes $d=\{512,1024,4096,16384\}$ and with CNN for sizes $d=\{512,1024,4096\}$.


For CNN-based architecture, $5$ hidden layers were used: $3$ convolutional $1$D layers followed by $2$ dense layers.
Due to its resemblance to the original LeNet architecture \cite{lecun1998gradient}, we refer to this architecture as to \textit{Text-LeNet}. 
The detailed configuration is provided in the Appendix. We noted in the experiments that Batch Normalization~\cite{DBLP:journals/corr/IoffeS15} is very helpful for faster convergence of the network and, therefore, it was added after every convolutional and dense layer.  The clip value for the gradients was set to $1$ during the backward pass.  RMSProp~\cite{hinton2012neural} was used as the optimizer with a learning rate of $1$e-$3$ and categorical cross-entropy as the loss function. All the checkpoints were saved in the h5 format. Larq Compute Engine~\cite{larq}, a highly optimised engine for quantization of networks, was used to convert the h5 files to tflite format for BNNs.

For classifiers, that were a part of the grid-based search, the paper reports the results for the best hyperparameters. 
For all other classifiers the default hyperparameter settings provided by the sklearn library were used. A $5$-fold cross-validation was used in the experiments. A total of $5$ simulations were performed and the average results are reported in the paper. All sklearn-based classifier experiments were performed on the CPU and CNN-based experiments were performed on NVIDIA Tesla GPUs.

\subsection{Results and Discussions}

To demonstrate the efficiency of binarized architecture for the intent classification task, we compare the results of binarized HD vectors with the binarized \textit{Text-LeNet} (BNN) architecture as classifier against non-binarized HD vectors with non-binarized \textit{Text-LeNet}. The $F_1$ scores are compared in Table~\ref{tab:classifiers:binarized} where BNN performed equally well to a \textit{Text-LeNet} architecture while being $20$\% to $40$\% more memory efficient, as shown in Figure~\ref{fig:memoryandtimeimage}(a). Due to the specifics of implementation, BNNs use $32$ bit float values as \textit{Text-LeNet}. The memory efficiency of BNNs can be further improved by $4$x when $8$-bit representations are used. On performance side, BNNs outperforms the \textit{Text-LeNet} for AskUbuntu and WebApplication datasets on $4$ out of $6$ tokenizers. The  results reported in Table~\ref{tab:classifiers:binarized} used $512$ dimensional HD vectors for Chatbot, AskUbuntu and WebApplication corpus, while $1024$ dimensional HD vectors were used for 20NewsGroups dataset. The choice of dimensionality  was based on the size of the datasets and a larger dimensionality was used for larger datasets. 

\begin{figure}[t]
\subfigure[Chatbot Corpus]{\includegraphics[width=0.5\linewidth]{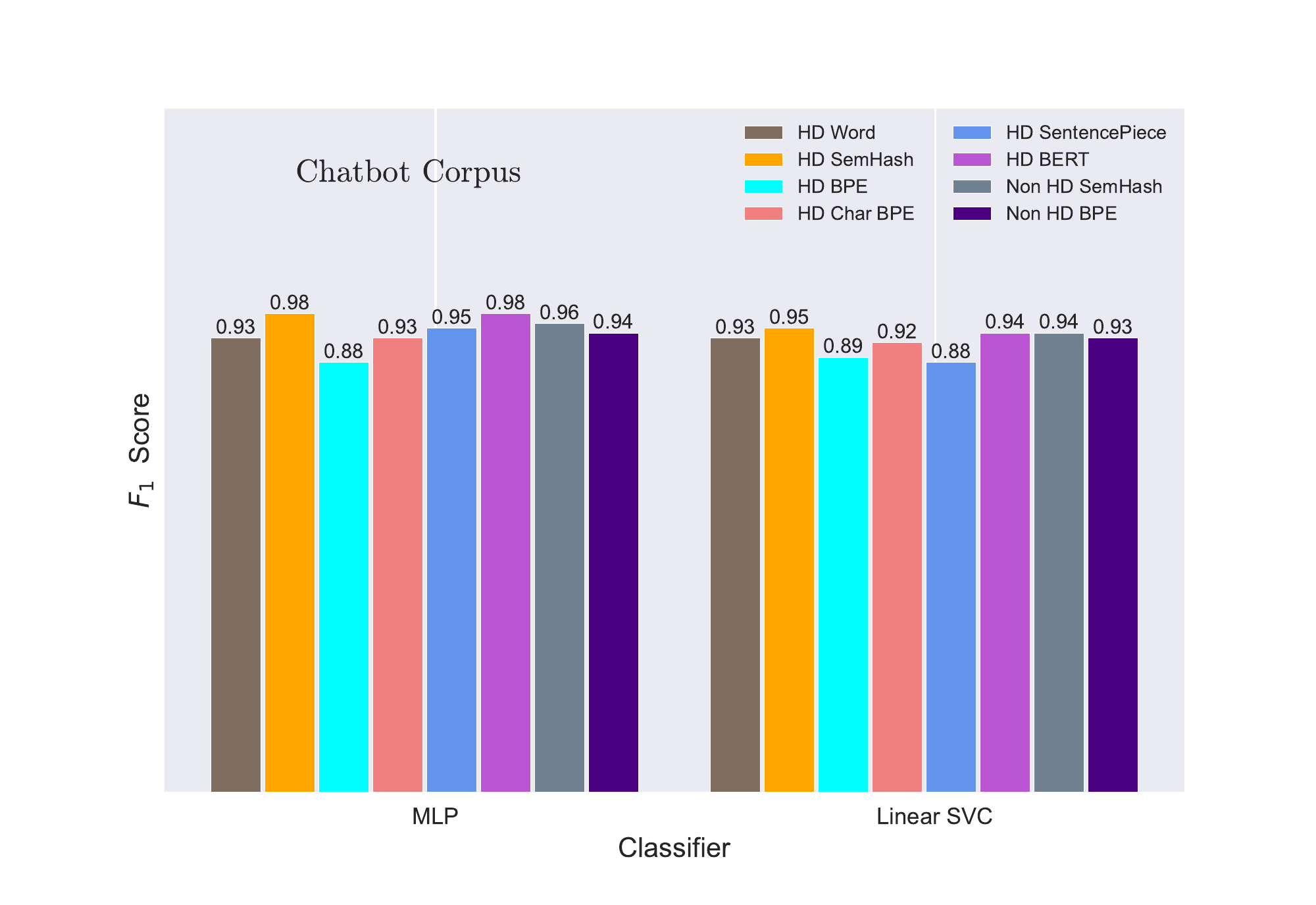}}\hfill
\subfigure[AskUbuntu Corpus]{\includegraphics[width=0.5\linewidth]{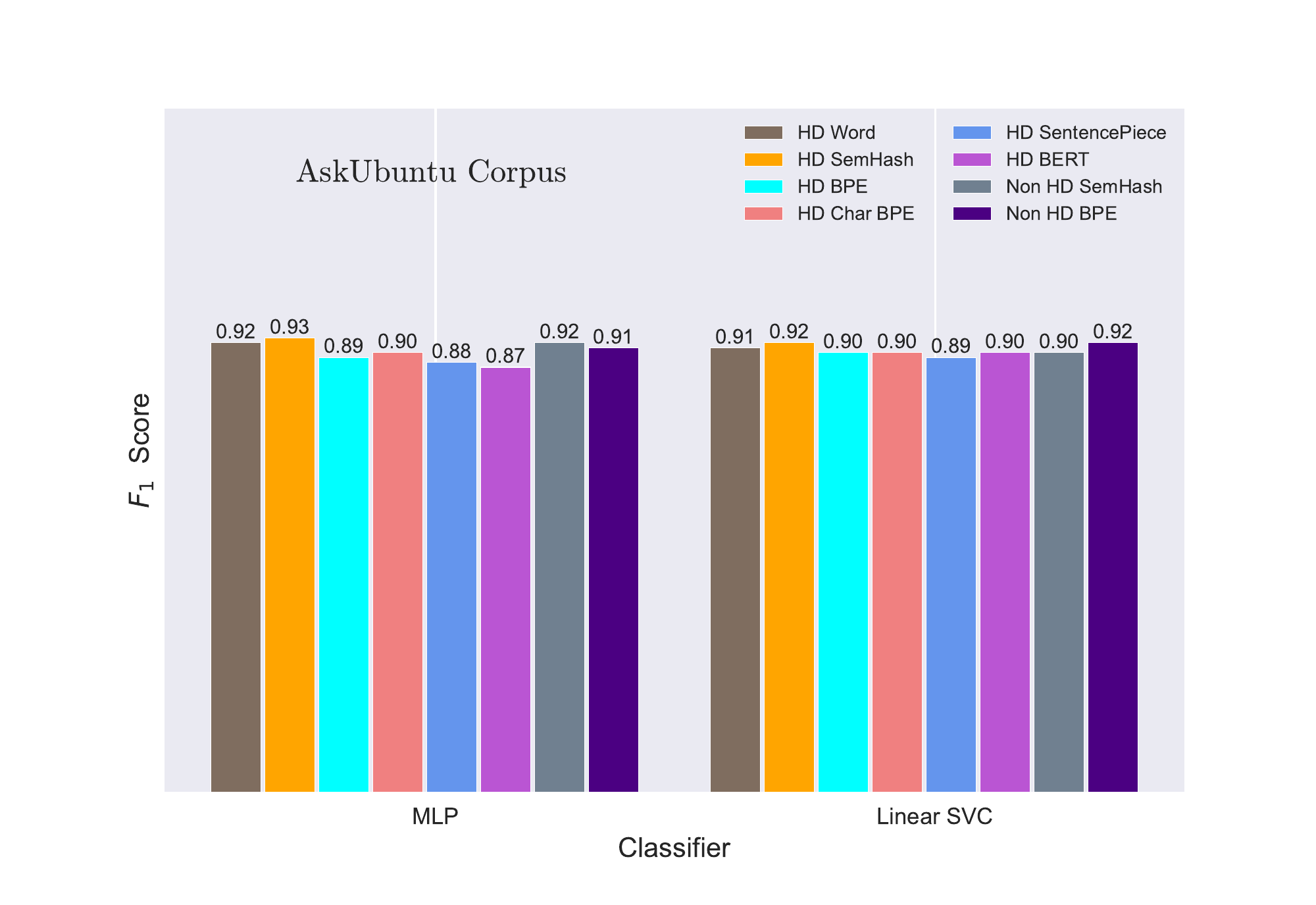}}\\[-2.4ex] 
\subfigure[WebApplication Corpus]{\includegraphics[width=0.5\linewidth]{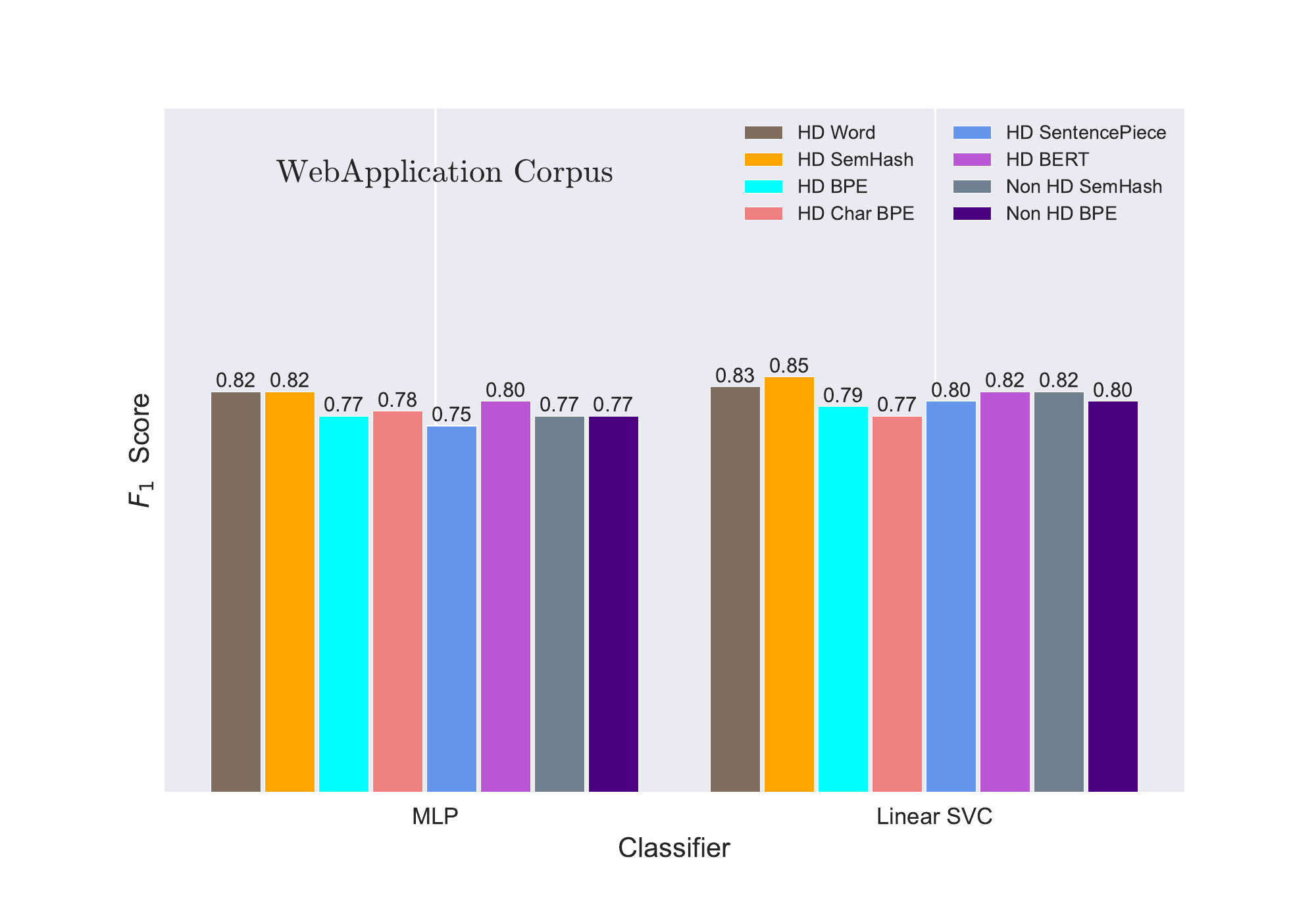}}\hfill
\subfigure[20NewsGroups Corpus]{\includegraphics[width=0.5\linewidth]{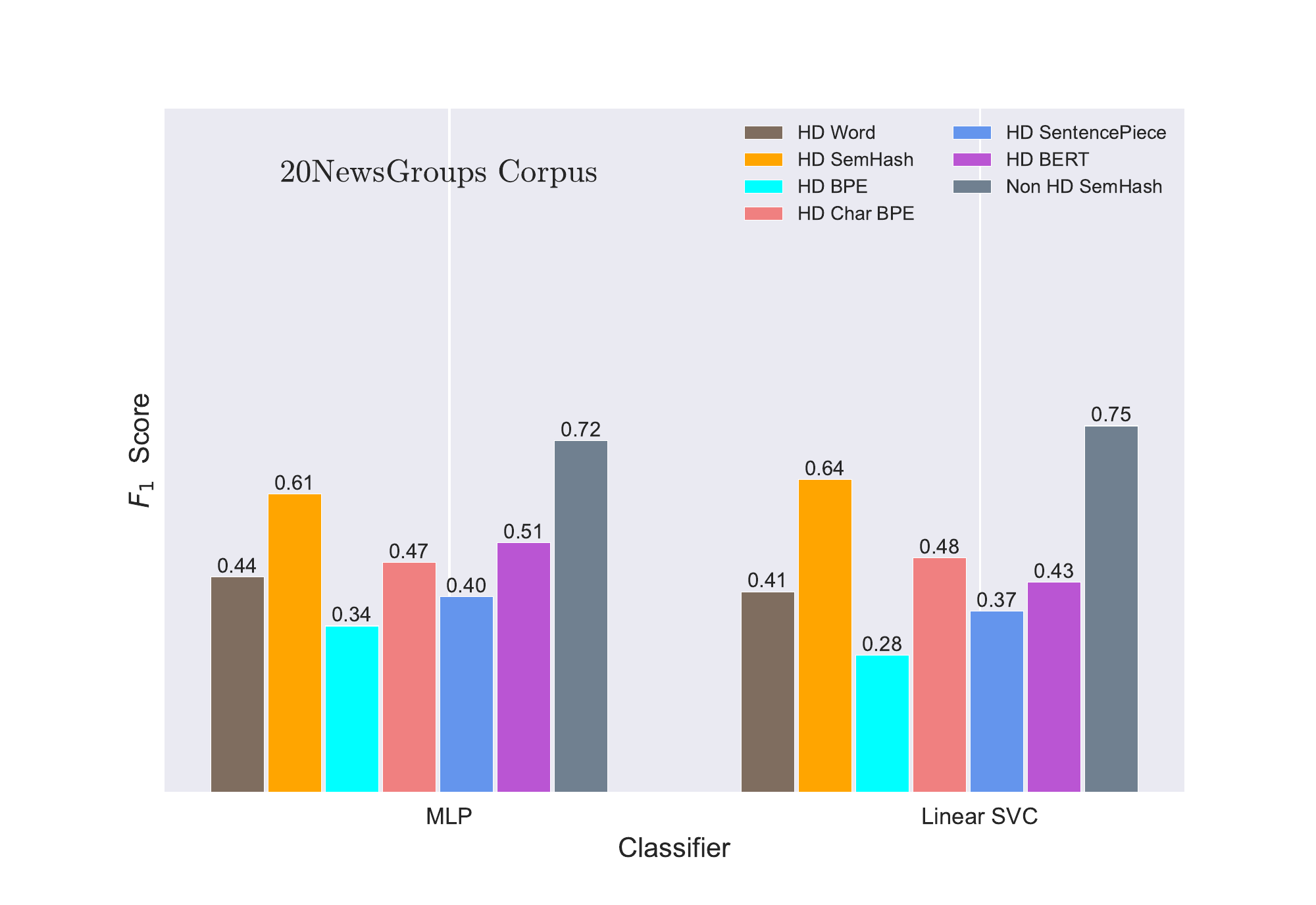}}
\caption{(a), (b), (c) and (d) shows the $F_1$ score comparison of MLP and Linear SVC classifier with HD and non-HD based tokenizers on Chatbot, AskUbuntu, WebApplication and 20NewsGroups corpus respectively.}
\label{fig:comparisonHDvsnon-HDChatbotAskUbuntu}
\end{figure}

One thing to note here is that \textit{Text-LeNet} also used HD vectors with the mentioned tokenizers, but the HD vectors were non-binarized. 
HD vectors in itself are already faster and much more efficient than counter-based representations, as shown in~\cite{alonso2020hyperembed}. 
When experimenting with other embedding methods like GloVe, the training was significantly slower; therefore,  HD vectors were used for all the experiments. In addition to that, using binarized classifier (BNN) further improves the training time upto $50\%$ compared to non-binarized classifier on all four datasets as shown in Figure ~\ref{fig:memoryandtimeimage} (b). Furthermore, when compared to GloVe embeddings with \textit{Text-LeNet}, HD BNN used around $20$\ -\ $40$\% lesser memory for all the intent classification datasets.

To justify that the binarization of HD vectors is effective and needed in this work, we benchmarked the binarized HD vectors with binarized $300$-dimensional GloVe vectors and the binarized version of counter-based representation for SemHash tokenizer for all the datasets. Table~\ref{tab:classifiers_BNN} summarizes the results of the comparison. All the binarized representations were trained with the same BNN classifier. Binarized HD vectors performed significantly better than other binarized methods outperforming binarized GloVe by $4$\ -\ $25$\% and binarized SemHash by $2$\ -\ $8$\% on $2$ out of $3$ smaller intent classification datasets and achieves comparable results for AskUbuntu dataset. The trend continued for 20NewsGroups with binarized HD achieving $5$\ -\ $7$\% better $F_1$ scores. Note that for the SemHash counter-based vectorizer, we put a sign function $sign(x) = +1\ \text{for}\ x>0\ \text{and}\ -1\ \text{otherwise}$.

In Figure~\ref{fig:comparisonHDvsnon-HDChatbotAskUbuntu}, MLP and Linear SVC with all the tokenizers with HD vectors as representation are compared with MLP and Linear SVC classifiers with SemHash tokenizers and counter-based vectorizer as representation from~\cite{alonso2020hyperembed}.  The $F_1$ score is comparable to the state-of-the-art for both MLP and SVC. For all small intent classification datasets, binarized HD vectors have achieved better results than non-HD vectors. The proposed architecture beats the non-HD baselines by $+2$\% for AskUbuntu and Chatbot Corpus, and $+5$\% for WebApplication Corpus. However, for 20NewsGroups the results of binarized HD Vectors are lower than non-HD Vectors. This is mainly due to the large size of the dataset and simple classifiers like LinearSVC failed to perform with just binarized values. The results for all the other classifiers are provided in the Appendix.  

Table~\ref{tab:f1 score comparison} compares the $F_1$ scores of various platforms on the intent classification datasets. We report the results of binarized HD vectors with the best classifiers from one of the nine classifiers mentioned (Binarized HD vectors with the best classifier), non-binarized HD vectors with \textit{Text-LeNet} (HD \textit{Text-LeNet}) and binarized HD vectors with binarized \textit{Text-LeNet} (HD BNN). Our end to end binarized architecture (HD BNN) achieved the state-of-the-art results for Chatbot dataset. 
The approach where only HD vectors were binarized (binarized HD vectors with the best classifier) achieved the state-of-the-art results for AskUbuntu dataset. 
The results on WebApplication dataset are comparable to the state-of-the-art ($0.87$ with SemHash): $0.84$ for binarized HD vectors with the best classifier and $0.83$ for HD BNN. 
The average performance of both  binarized HD vectors with the best classifier ($0.92$) and HD BNN ($0.91$) was also comparable to the best non-binarized approach ($0.92$ with Semhash).  

\begin{table}[t]

\begin{center}
\scalebox{0.85}{
\begin{tabular}{|c|c|c|c|c|} 
\hline
Platform & Chatbot & AskUbuntu & WebApplication & Average \\ [1ex] 
\hline
Botfuel  &  0.98 & 0.90 & 0.80 & 0.89\\ 
\hdashline     
Luis & 0.98 & 0.90 & 0.81 & 0.90\\ 
\hdashline      
Dialogflow  & 0.93 & 0.85 & 0.80 & 0.86\\
\hdashline      
Watson &  0.97 & 0.92 & 0.83 & 0.91\\
\hdashline      
Rasa  &  0.98 & 0.86 & 0.74 & 0.86\\
\hdashline      
Snips &  0.96 & 0.83 & 0.78  & 0.86\\ 
\hdashline      
Recast & \textbf{0.99} & 0.86 & 0.75 & 0.87\\
\hdashline
TildeCNN & \textbf{0.99} & 0.92 & 0.81  & 0.91\\
\hdashline
FastText & 0.97 & 0.91 & 0.76 & 0.88\\
\hdashline
SemHash & 0.96 & 0.92 & \textbf{0.87} & \textbf{0.92}\\
\hdashline
BPE & 0.95 & \textbf{0.93} & 0.85  & 0.91\\
\hdashline
HD vectors ~\cite{alonso2020hyperembed} & 0.97 & 0.92 & 0.82  & 0.90\\
\hline \hline
Binarized HD vectors with the best classifier & 0.98 & \textbf{0.93} & 0.84  & \textbf{0.92} \\
HD \textit{Text-LeNet} & 0.94 & 0.87 & 0.79  & 0.88 \\
HD BNN& \textbf{0.99} & 0.84 & 0.83  & 0.91 \\
\hline
\end{tabular} }

\caption{$F_1$ score comparison of various platforms on intent classification datasets of short texts with methods used in the paper. Some results are taken from~\cite{alonso2020hyperembed}}
\label{tab:f1 score comparison}
\end{center}
\end{table}

\section{Conclusion}

In this work, we present an end to end binarized architecture for the intent classification task.
We demonstrate that it is possible to achieve comparable to the state-of-the-art results while using the binarized representations of all the components of the NLP classification architecture. 
This allows exploring the effectiveness of binary representations both for reducing the memory footprint of the architecture and for increasing the energy-efficiency of the inference phase due to effectiveness of binary operations. 
This work takes a step towards enabling NLP functionality on resource-constrained devices.

\section*{Acknowledgement}

The authors would like to thank P.~Alonso, E.~Osipov, and M.~Liwicki with whom the previous research, which is the foundation of this work, has been done. 
The work of DK was supported by the European Union’s Horizon 2020 Research and Innovation Programme under the Marie Skłodowska-Curie Individual Fellowship Grant Agreement 839179 and in part by the DARPA’s VIP (Super-HD Project) and AIE (HyDDENN Project) programs.

\bibliographystyle{coling}
\bibliography{coling2020}

\newpage

\appendix
\section{Appendix}

\setcounter{table}{0}

\subsection{Datasets}
The experiments in the paper were performed on four datasets: the \textit{Chatbot Corpus} (Chatbot), the \textit{Ask Ubuntu Corpus} (AskUbuntu), the \textit{Web Applications Corpus} (WebApplication), and the \textit{20 News Groups Corpus} (20NewsGroups).

A chatbot was created on Telegram, where questions of the public transport of Munich were posted. The chatbot replied to the questions and thus data was collected for the Chatbot corpus. A detailed test and train split are provided in Table \ref{tab:data distribution Chatbot}. 

\begin{table}[ht]
\begin{center}
\scalebox{0.90}{
\begin{tabular}{|c | c | c| c|} 
\hline
Intent & Train original & Train Augmented & Test  \\ [0.75ex] 
\hline
Departure Time &  43 & 57 & 35 \\ 
Find Connection & 57 & 57 & 71 \\
\hline
\end{tabular} }
\caption{Data sample distribution for the Chatbot dataset}
\label{tab:data distribution Chatbot}
\end{center}
\end{table}

The AskUbuntu and WebApplication datasets are questions and answers from the StackExchange platform. A detailed breakdown is provided in Table \ref{tab:data distribution AskUbuntu} and \ref{tab:data distribution WebApplication} respectively.  

\begin{table}[ht]

\begin{center}
\scalebox{0.90}{
\begin{tabular}{|c | c | c| c|} 
\hline
Intent & Train original & Train Augmented & Test \\ [1ex] 
\hline
Make Update &  10 & 17 & 37  \\ 
 
Setup Printer &  10 & 17 & 13 \\ 
     
Shutdown Computer &  13 & 17 & 14 \\ 
     
Software Recommendation &  17 & 17 & 40 \\ 
     
None & 3 & 17  & 5 \\

\hline
\end{tabular}}
\caption{Data sample distribution for the AskUbuntu dataset}
\label{tab:data distribution AskUbuntu}
\end{center}
\end{table}

\begin{table}[ht]

\begin{center}
\scalebox{0.90}{
\begin{tabular}{|c | c | c| c|}  
\hline
Intent & Train original & Train Augmented & Test \\ [1ex] 
\hline
Change Password &  2 & 7 & 6 \\ 
     
Delete Account &  7 & 7 & 10 \\ 
     
Download Video &  1 & 7 & 0 \\
     
Export Data &  2 & 7 & 3 \\
     
Filter Spam &  6 & 7 & 14 \\
     
Find Alternative &  7 & 7 & 16 \\ 
     
Sync Accounts &  3 & 7 & 6 \\ 
     
None & 2 & 7 & 4 \\

\hline
\end{tabular} }

\caption{Data sample distribution for the WebApplication dataset}
\label{tab:data distribution WebApplication}
\end{center}
\end{table}

The 20NewsGroups dataset comprises news posts labelled into several categories and a detailed breakdown is provided in Table \ref{tab:data:distribution:20newsgroups}.

\begin{table}[ht]

\begin{center}
\scalebox{0.90}{
\begin{tabular}{|c | c | c|} 
\hline
Categories & Train & Test \\ [1ex] 
\hline
alt.atheism &  11314 & 7532\\ 
     
comp.graphics &  11314 & 7532\\ 
     
comp.os.ms-windows.misc &  11314 & 7532\\
     
comp.sys.ibm.pc.hardware &  11314 & 7532\\
     
comp.sys.mac.hardware &  11314 & 7532\\
     
comp.windows.x &  11314 & 7532\\
     
misc.forsale  &  11314 & 7532\\
     
rec.autos &  11314 & 7532\\

rec.motorcycles &  11314 & 7532\\

rec.sport.baseball &  11314 & 7532\\

rec.sport.hockey &  11314 & 7532\\

sci.crypt &  11314 & 7532\\

sci.electronics &  11314 & 7532\\

sci.electronics &  11314 & 7532\\

sci.space &  11314 & 7532\\

soc.religion.christian &  11314 & 7532\\

talk.politics.guns &  11314 & 7532\\

talk.politics.mideast &  11314 & 7532\\

talk.politics.misc &  11314 & 7532\\

talk.religion.misc &  11314 & 7532\\


\hline
\end{tabular} }

\caption{Data sample distribution for the 20NewsGroups dataset}
\label{tab:data:distribution:20newsgroups}
\end{center}
\end{table}

\subsection{Experimentation Details}

The \textit{Text-LeNet} architecture used in the experiments is defined as follows:

\paragraph{Text-LeNet}: $[128, M, BN, 256, M, BN, 512, M, F, 128_D, DO, C]$\\

\noindent where, numbers 128, 256 and 512 represents the filters of \emph{Convolution layer} which is followed by an activation function. \emph{M} represents the \emph{Max Pooling layer} and \emph{BN} represents the \emph{Batch Normalization layer}. \emph{F} refers to a \emph{Flatten layer}. $128_D$ represents the \emph{Dense layer} of size $128$ followed by a \emph{Dropout layer} denoted by \emph{DO}. Finally, \emph{C} represents the \emph{Linear classification layer} of dimension ($128$, number of classes).

\newpage

\noindent Other hyper-parameters settings include:

\begin{table}[!h]

  \centering
  \begin{tabular}{ll}
    \toprule                
    \textbf{Hyper-parameter} & \textbf{Value}  \\
    \midrule
    Convolution Kernel Size & $3$     \\
    Convolution layer Padding & valid  \\
    Max-Pooling Kernel Size & $3$,$2$,$3$ for the three $M$ layers respectively    \\
    Optimizer & RMSprop \\
    Loss & Categorical cross entropy\\
    Activation Function & Rectified Linear Unit (ReLU)\\
    Batch Size & $4$ (small datasets) and   $64$ (20NewsGroups)  \\
    Learning Rate & $0.001$ \\
    Number of Epochs & $15$-$40$ (Depending upon the dataset)  \\
    Initializer & Xavier initialization\\
    \bottomrule
  \end{tabular}
  \caption{Hyper-parameters for the experiments}
    \label{hyper-parameters}
\end{table}

To find the best hyperparameters, a grid-based search was applied to all the datasets for the MLP classifier. Three different hidden layer configurations were considered:  [($100,50$), ($300,100$), ($300,200,100$)]. The best parameters were chosen based on \texttt{GridSearchCV}. The maximal number of iterations for MLP  was set to $200$.

In the case of Random Forest, two hyperparameters include optimized number of estimators ([$50, 60, 70$]) and minimum samples leaf ([$1, 11$]); we used  $50$ estimators and $1$ leaf.
In the case of KNN, the number of neighbors between $3$ and $7$ was considered; $3$ neighbors were used in the experiments.  
For all the other classifiers the default hyperparameter settings provided by the sklearn library were used.

\newpage

\subsection{Experiments}

Table \ref{tab:WebApplication f1 Scores}-\ref{tab:Chatbot f1 Scores} compares the $F_1$ scores of nine classifiers: MLP, Random Forest, Linear SVC, Passive Aggressive, SGD Classifier, Ridge Classifier, Nearest Centroid, Bernoulli NB, KNN Classifier on all three small datasets with HD versions of six tokenization methods and Non HD versions of SemHash and BPE for small datasets and Non HD SemHash for 20NewsGroups Corpus. SP is an acronym for Sentence Piece. Table \ref{tab:20NewsGroups f1 Scores} compares the results of all the tokenizers using four classifiers: MLP, SGD Classifier, Linear SVC and Random Forest for the 20NewsGroups dataset. SemHash tokenizer, in general, achieved better results compared to other tokenizers followed by BERT tokenizer on all four datasets. 

\begin{table}[h]
\begin{center}

\scalebox{0.75}{
\begin{tabular}{|c|c|c|c|c|c|c|c|c|} 
\hline
Classifier & HD Word & HD SemHash & HD BPE & HD Char BPE & HD SP & HD BERT & Non HD SemHash & Non HD BPE  \\ [1ex] 
\hline
MLP  &  0.92 & \textbf{0.93} & 0.89 & 0.90 & 0.88 & 0.87 & 0.92 & 0.91\\ 
\hdashline  
Passive Aggr. & 0.92 & \textbf{0.93} & 0.90 & 0.88 & 0.91 & 0.90 & 0.92 & \textbf{0.93}\\ 
\hdashline      
SGD Classifier  & 0.86 & \textbf{0.89} & 0.85 & 0.88 & 0.86 & \textbf{0.89} & \textbf{0.89} & \textbf{0.89} \\
\hdashline     
Ridge Classifier &  \textbf{0.92} & \textbf{0.92} & 0.88 & 0.87 & 0.91 & \textbf{0.92} & 0.90 & 0.91  \\
\hdashline    
KNN Classifier  &  0.81 & \textbf{0.84} & 0.78 & 0.60 & 0.80 & 0.79 & 0.79 & 0.72 \\
\hdashline      
Nearest Centroid &  \textbf{0.91} & \textbf{0.91} & 0.88  & 0.86 & 0.86 & 0.85 & 0.90 & 0.89 \\ 
\hdashline  
Linear SVC & 0.91 & \textbf{0.92} & 0.90 & 0.90 & 0.89 & 0.90 & 0.90 & \textbf{0.92} \\
\hdashline
Random Forest & 0.90 & \textbf{0.92} & 0.82  & 0.83 & 0.83 & 0.82 & 0.88 & 0.90 \\
\hdashline
Bernoulli NB & 0.76 & 0.87 & 0.74 & 0.84 & 0.65 & 0.79 & 0.91 & \textbf{0.92} \\
\hline

\hline
\end{tabular}}
\caption{$F_1$ scores of all sklearn classifiers for AskUbuntu dataset. }
\label{tab:WebApplication f1 Scores}
\end{center}
\end{table}

\begin{table}[h]
\begin{center}
\scalebox{0.75}{
\begin{tabular}{|c|c|c|c|c|c|c|c|c|} 
\hline
Classifier & HD Word & HD SemHash & HD BPE & HD Char BPE & HD SP & HD BERT & Non HD SemHash & Non HD BPE  \\ [1ex]
\hline
MLP  &  0.93 & \textbf{0.98} & 0.88 & 0.93 & 0.95 & \textbf{0.98} & 0.96 & 0.94 \\ 
\hdashline     
Passive Aggr. & \textbf{0.96} & \textbf{0.96} & 0.92 & 0.94 & 0.89 & 0.94 & 0.95 & 0.91\\ 
\hdashline      
SGD Classifier  & 0.95 & \textbf{0.97} & 0.93 & 0.89 & 0.89 & 0.92 & 0.93 & 0.93\\
\hdashline      
Ridge Classifier &  0.94 & \textbf{0.95} & 0.77 & 0.91 & 0.84 & 0.88 & 0.94 & 0.94\\
\hdashline      
KNN Classifier  &  0.86 & 0.86 & 0.80 & 0.83 & 0.80 & \textbf{0.91} & 0.75 & 0.71 \\
\hdashline      
Nearest Centroid &  0.79 & 0.89 & 0.90  & 0.86 & 0.88 & 0.88 & 0.89 & \textbf{0.94} \\ 
\hdashline   
Linear SVC & 0.93 & \textbf{0.95} & 0.89 & 0.92 & 0.88 & 0.94 & 0.94 & 0.93\\
\hdashline
Random Forest & 0.88 & 0.91 & 0.77  & 0.83 & 0.85 & 0.92 & \textbf{0.95} & \textbf{0.95} \\
\hdashline
Bernoulli NB & 0.81 & 0.88 & 0.81 & 0.89 & 0.84 & 0.91 & \textbf{0.93} & \textbf{0.93}\\
\hline

\hline
\end{tabular} }
\caption{$F_1$ scores of all sklearn classifiers for Chatbot dataset. }
\label{tab:AskUbuntu f1 Scores}
\end{center}
\end{table}

\begin{table}[h]
\begin{center}
\scalebox{0.75}{
\begin{tabular}{|c|c|c|c|c|c|c|c|c|} 
\hline
Classifier & HD Word & HD SemHash & HD BPE & HD Char BPE & HD SP & HD BERT & Non HD SemHash & Non HD BPE  \\ [1ex]
\hline
MLP  &  \textbf{0.82} & \textbf{0.82} & 0.77 & 0.78 & 0.75 & 0.80 & 0.77 & 0.77\\ 
\hdashline     
Passive Aggr. & \textbf{0.83} & \textbf{0.83} & 0.76 & 0.80 & 0.80 & 0.80 & 0.82 & 0.80 \\ 
\hdashline      
SGD Classifier  & 0.76 & 0.79 & 0.66 & 0.64 & 0.64 & \textbf{0.84} & 0.75 & 0.74\\
\hdashline      
Ridge Classifier &  0.79 & \textbf{0.84} & 0.79 & 0.78 & 0.76 & 0.81 & 0.79 & 0.80\\
\hdashline      
KNN Classifier  &  0.80 & 0.81 & 0.78 & 0.67 & 0.80 & \textbf{0.84} & 0.72 & 0.75 \\
\hdashline      
Nearest Centroid &  0.77 & \textbf{0.80} & 0.72  & 0.76 & 0.75 & 0.75 & 0.74 & 0.73\\ 
\hdashline   
Linear SVC & 0.83 & \textbf{0.85} & 0.79 & 0.77 & 0.80 & 0.82 & 0.82 & 0.80\\
\hdashline
Random Forest & 0.77 & 0.80 & 0.61  & 0.68 & 0.67 & 0.78 & \textbf{0.87} & 0.85 \\
\hdashline
Bernoulli NB & 0.61 & 0.70 & 0.57 & 0.62 & 0.57 & 0.61 & 0.74 & \textbf{0.75} \\
\hline

\hline
\end{tabular} }

\caption{$F_1$ scores of all sklearn classifiers for WebApplication dataset. }
\label{tab:Chatbot f1 Scores}
\end{center}
\end{table}

\begin{table}[!htbp]
\begin{center}
\scalebox{0.75}{
\begin{tabular}{|c|c|c|c|c|c|c|c|} 
\hline
Classifier & HD Word & HD SemHash & HD BPE & HD Char BPE & HD SP & HD BERT & Non HD SemHash  \\ [1ex]
\hline
MLP  &  0.44 & 0.61 & 0.34 & 0.47 & 0.40 & 0.51 & \textbf{0.72}\\ 
\hdashline     
SGD Classifier  & 0.43 & 0.59 & 0.41 & 0.26 & 0.45 & 0.49 & \textbf{0.70}\\
\hdashline
Linear SVC  & 0.41 & 0.64 & 0.28 & 0.48 & 0.37 & 0.43 & \textbf{0.75}\\
\hdashline
Random Forest & 0.17 & 0.33 & 0.13  & 0.20 & 0.13 & 0.21 & \textbf{0.58}\\

\hline

\hline
\end{tabular} }

\caption{$F_1$ scores of sklearn classifiers for 20NewsGroups dataset. }
\label{tab:20NewsGroups f1 Scores}
\end{center}
\end{table}

\end{document}